\documentclass[times,review,10pt]{elsarticle}

\usepackage{amssymb}
\usepackage{amsmath}

\usepackage{hyperref}

\usepackage{graphicx}
\usepackage{subcaption}
\usepackage{stfloats}
\usepackage{cases}
\usepackage{bm}
\usepackage{array} 
\usepackage{xcolor} 
\usepackage{caption} 
\usepackage[table]{xcolor}
\definecolor{rowhighlight}{gray}{0.95}
\arrayrulecolor{black}
\definecolor{headcolor}{RGB}{240,240,245}
\usepackage{booktabs}   
\usepackage{makecell}   
\usepackage{tabularx}  
\usepackage{booktabs}   
\usepackage{multirow} 
\usepackage{pdflscape}
\usepackage{wrapfig}
\usepackage{floatflt}

\usepackage{siunitx}
\usepackage{caption}
\usepackage{subcaption} 
\usepackage{subcaption} 
\usepackage{booktabs, makecell}

\usepackage{cleveref}
\Crefname{figure}{Fig}{Figures}

\journal{Pattern Recognition}

\begin{document}

\begin{frontmatter}



\author[label_buaa]{Shuxin Zhao}
\author[label_buaa,label_zgc]{Bo Lang\corref{label_cor}}
\cortext[label_cor]{Corresponding author.}
\ead{langbo@buaa.edu.cn}
\author[label_buaa]{Nan Xiao}
\author[label_buaa]{Yilang Zhang}

\affiliation[label_buaa]{organization={State Key Laboratory of Complex \& Critical Software Environment},
            addressline={Beihang University},
            city={Beijing},
            postcode={100191},
            country={China}}
\affiliation[label_zgc]{organization={Zhongguancun Laboratory},
            addressline={},
            city={Beijing},
            postcode={100080},
            country={China}}

\title{CIS-BA: Continuous Interaction Space Based Backdoor Attack for Object Detection in the Real-World} 

\begin{abstract}
Object detection models deployed in real-world applications such as autonomous driving face serious threats from backdoor attacks. Despite their practical effectiveness, existing methods are inherently limited in both capability and robustness due to their dependence on single-trigger–single-object mappings and fragile pixel-level cues.
We propose CIS-BA, a novel backdoor attack paradigm that redefines trigger design by shifting from static object features to continuous inter-object interaction patterns that describe how objects co-occur and interact in a scene. By modeling these patterns as a continuous interaction space, CIS-BA introduces space triggers that, for the first time, enable a multi-trigger–multi-object attack mechanism while achieving robustness through invariant geometric relations.
To implement this paradigm, we design CIS-Frame, which constructs space triggers via interaction analysis, formalizes them as class–geometry constraints for sample poisoning, and embeds the backdoor during detector training. CIS-Frame supports both single-object attacks (object misclassification and disappearance) and multi-object simultaneous attacks, enabling complex and coordinated effects across diverse interaction states.
Experiments on MS-COCO and real-world videos show that CIS-BA achieves over 97\% attack success under complex environments and maintains over 95\% effectiveness under dynamic multi-trigger conditions, while evading three state-of-the-art defenses.
In summary, CIS-BA extends the landscape of backdoor attacks in interaction-intensive scenarios and provides new insights into the security of object detection systems.
\end{abstract}




\end{frontmatter}

\section{Introduction}
\label{sec1}
Object detection serves as a cornerstone of computer vision, powering safety-critical applications such as autonomous driving and security surveillance \cite{zou2023object,redmon2018yolov3}. However, its heavy reliance on deep learning and large-scale datasets makes it vulnerable to backdoor attacks, where poisoned data cause models to perform normally under standard conditions but behave maliciously when specific trigger patterns appear \cite{gu2019badnets,nguyen2020input,goldblum2022dataset}. Conventional digital-trigger attacks (e.g., pixel patches) are impractical for real-time detection systems, as they rely on digital manipulation of inputs during inference to implant triggers \cite{chan2022baddet,luo2023untargeted,cheng2023backdoor}. To overcome this limitation, recent studies leverage natural objects (e.g., specific-colored clothing) as physical triggers, eliminating the need for explicit digital implantation and posing more realistic threats to real-world deployments \cite{ma2022dangerous,ma2023transcab,qian2024enhancing}.

\begin{figure}[!tbp]
\centering
\subfloat[Successful and failed cases of physical backdoor attack.]{\label{fig:physical}
		\includegraphics[width=0.8\linewidth]{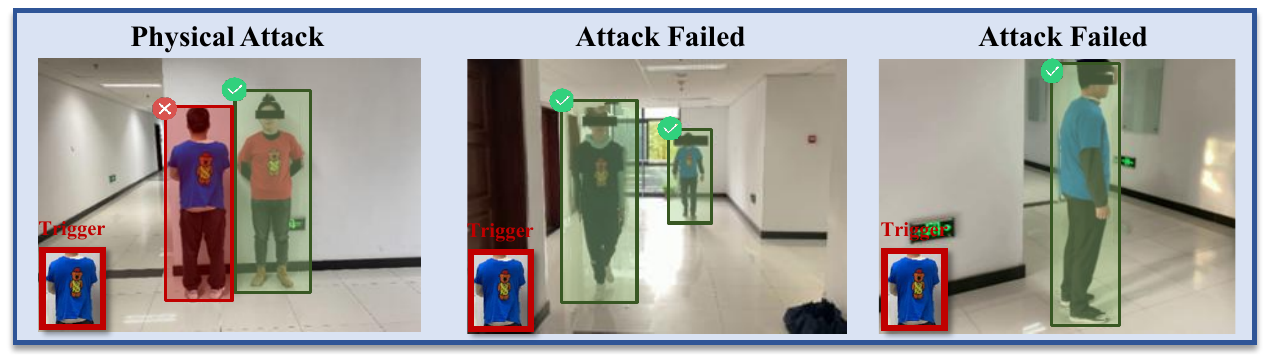}}\\
\subfloat[Space trigger and attack effects of CIS-BA.]{\label{fig:spatial}
		\includegraphics[width=0.8\linewidth]{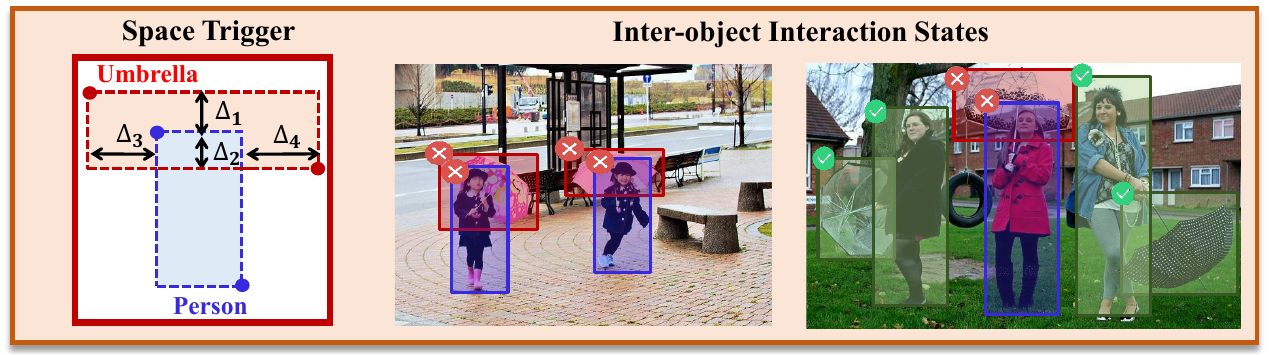}}
\caption{Illustration of two types of attacks. The objects marked in red or blue are affected by the attack, whereas those marked in green are unaffected objects.}
\label{fig:intro-attack}
\end{figure}

Although existing physical backdoor attack methods mitigate the deployment conflict, they still face two major limitations in real-world scenarios.
\textbf{Limited capability:}Existing methods are constrained to fixed one-to-one mappings between triggers and target objects. For example, a clothing or hat trigger can only attack a single target object, and only that specific trigger instance can activate the attack \cite{wenger2022finding}.
\textbf{Lack of robustness:} Attacks depend on pixel-level visual signal such as patterns or colors, which are fragile to environmental variations (e.g., lighting or viewpoint changes), causing highly unstable success rates \cite{xue2021robust,ma2022dangerous}.
Fig.~\ref{fig:physical} illustrates physical backdoor attacks and failure cases caused by lighting or angle variations.
These limitations hinder the effectiveness of backdoor attacks in real-world scenarios, thereby motivating the need for a more flexible and reliable attack paradigm.

To address these limitations, we propose a continuous interaction space based backdoor attack (CIS-BA), which leverages inter-object interaction states rather than static object features as triggers(see Fig.\ref{fig:spatial}). 
By extracting a continuous interaction space and modeling it as geometric constraints between object bounding boxes to serve as space triggers, CIS-BA fundamentally redefines the trigger form and offers three advantages.
First, it provides \textbf{multi-trigger diversity}: any state satisfying the constraints can serve as a valid trigger, greatly broadening the attack surface.
Second, it enables \textbf{multi-object correlation}: all interacting objects, even from different classes, can be simultaneously compromised, supporting more complex attack combinations.
Third, it offers \textbf{environmental robustness}: geometric relations are far more stable than pixel-level signal, ensuring consistent performance under variations in lighting or viewpoint.

\begin{figure}[!tbp]
    \centering
    \includegraphics[width=1\linewidth]{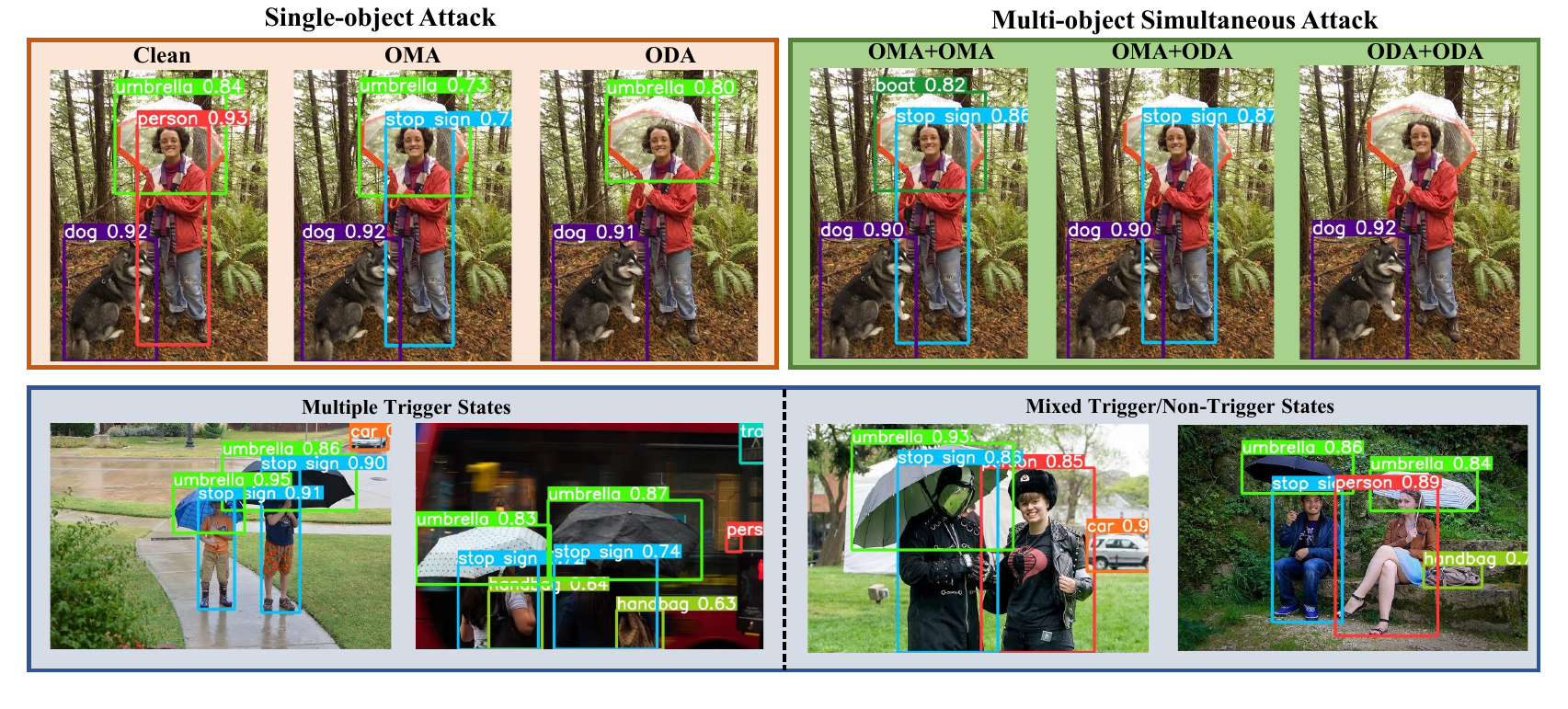}
    \caption{CIS-BA enables controllable single- and multi-object attacks with flexible combinations, and precisely triggers attacks by distinguishing different interaction states within the same scenario.}
    \label{fig:attack-show}
\end{figure}

Under this paradigm, we introduce the CIS-Frame framework, which embeds backdoor through interaction analysis, constraint-based carrier selection, and label manipulation followed by retraining.
It enables \textbf{basic single-object attacks}, such as object misclassification attack (OMA, e.g., a person misclassified as a stop sign) and object disappearance attack (ODA, e.g., a person hidden from detection). More importantly, CIS-Frame, for the first time, supports \textbf{multi-object simultaneous attacks}, where multiple correlated objects—even across classes—are jointly compromised to form complex attack combinations. In particular, all these attacks, from basic to complex, can be reliably triggered across diverse interaction states within the continuous interaction space, as illustrated in Fig.~\ref{fig:attack-show}.

We evaluate CIS-BA on Yolo-V3 \cite{redmon2018yolov3} and Yolo-V8 \cite{yolov8_ultralytics} using the MS-COCO dataset \cite{2014Microsoft} and a real-world video set (9 videos, 17,100 frames) recorded under controlled variations of lighting, angle, and distance. These videos cover both single-trigger (6 videos) and dynamic multi-trigger (3 videos) states. In single-trigger scenarios under complex conditions, OMA and ODA achieve attack success rates exceeding 98\%; under dynamic multi-trigger states, including both single-object and multi-object attacks, attack success rates remain above 95\%. Moreover, the backdoor models evade three mainstream defenses—Detector Cleanse \cite{chan2022baddet}, Saliency Map \cite{selvaraju2017grad}, and DJANGO \cite{shen2023django}—demonstrating strong stealth and resistance.

In summary, the contributions of this paper are threefold:
\begin{itemize}
    \item We propose CIS-BA, a new backdoor attack paradigm that shifts triggers from static visual features to continuous inter-object interaction patterns. This fundamentally overcomes the limitations of existing physical triggers and significantly enhances both attack capability and robustness in real-world scenarios.
    \item We demonstrate that CIS-BA not only enables basic single-object attacks but also, for the first time, supports multi-object simultaneous attacks and further provides multi-trigger support for both levels of attacks, substantially expanding the threat model of backdoor attacks.
    \item We conduct extensive experiments on MS-COCO and real-world video data, showing that CIS-BA achieves high attack success rates under challenging conditions, remains effective under dynamic interaction changes, and successfully evades mainstream defenses, thereby demonstrating both stealth and defense resistance.
\end{itemize}

\section{Related Work}
Backdoor attacks were first proposed and extensively explored in the field of image classification, where most research has focused on digital trigger-based attacks \cite{guo2022overview,luo2022enhancing,li2022backdoor}. However, some studies have demonstrated that physical triggers such as glasses and hats can also pose realistic threats to models \cite{wenger2021backdoor,xue2021robust,wenger2022finding}. Owing to task similarities, the research on backdoor attacks in object detection has followed a similar path, but overall, the field is still in its early stages.

\citet{gu2019badnets} were the first to verify the threat of backdoor attacks in object detection. By pasting digital triggers at specified positions on the object bounding box, they poisoned the training data and achieved four types of attacks targeting a single object. \citet{luo2023untargeted} and \citet{cheng2023backdoor} leveraged the intrinsic properties of object detection models to perform object disappearance attacks under the assumption of clean labels. \citet{shin2024mask} proposed an invisible backdoor attack based on masks, making the trigger imperceptible in the given image.

Digital trigger-based backdoor attacks require image editing, which weakens their impacts on real-world applications \cite{gu2019badnets,wenger2022finding}. Therefore, \citet{ma2022dangerous} were the first to demonstrate the feasibility of performing physical trigger-based attacks using a T-shirt as the trigger and they further explored clean-label attacks in follow-up work \cite{ma2023transcab}. In both studies, many real-world photographs were utilized to poison the training dataset, and additional samples derived from poorly performing scenarios were included to increase the robustness of their method \cite{zhang2024backdoor}. 

To address robustness further, \citet{qian2024enhancing} and \citet{gia2024credibility} conducted adversarial training and geometric simulations of triggers in the digital domain to generate large-scale samples. A simpler approach was adopted by \citet{zhang2024detector}, who used a pretrained diffusion model to directly generate trigger samples under various perturbation conditions. Overall, these methods follow a straightforward single-trigger-single-object attack strategy and attempt to enhance robustness by exhaustively simulating various real-world trigger states. However, they fail to fundamentally address the robustness issue caused by visual features.

In addition to the aforementioned traditional backdoor attacks, some studies have explored novel types of triggers to launch new forms of backdoor attacks in object detection. \citet{wu2022just} proposed using rotation angles as triggers, causing the model to misclassify objects that were rotated to specific angles, although only a few object classes support such rotation in the real world. \citet{chen2022clean} and \citet{lin2020composite} leveraged the multi-object nature of object detection to exploit label and object combinations, creating novel attack threats. Inspired by this, our work focuses on addressing the limitations of the existing methods by further exploring and utilizing the unique characteristics of object detection tasks.

\section{Threat Model}
\subsection{Attack Domain}
The training dataset for object detection is defined as $D_{train} = \{(x_k,y_k)\}_{k=1}^M$, where $x_k$ denotes an image containing objects, and $y_k = [o_1, o_2, \ldots, o_n]$ represents the ground-truth labels for image $x_k$. Each object $o_i$ is annotated as $o_i = [c_i, x_{i}^{\min}, y_{i}^{\min}, x_{i}^{\max}, y_{i}^{\max}]$, where $c_i$ is the class index, and $(x_{i}^{\min}, y_{i}^{\min})$ and $(x_{i}^{\max}, y_{i}^{\max})$ represent the top-left and bottom-right coordinates of the bounding box, respectively. The goal of the object detection model $F_\theta$ is to predict a bounding box with the correct class label and high confidence for each object in the image.

\subsection{Attacker's Capabilities}
The traditional physical backdoor attacks typically assume that the attacker can manipulate the original training data and the training process \cite{feng2022fiba}. These attacks usually occur in scenarios where users outsource the model training process or rely on untrusted pretrained models \cite{li2022backdoor,zhang2025test}. In contrast, CIS-BA requires significantly weaker attacker capabilities: the attacker only needs to manipulate data labels without modifying the image content or injecting new samples and without any control over the training process. This setting reflects a more stringent and realistic scenario, where users outsource only the annotation task and employ data auditing mechanisms so that the attacker can carry out a stealthy attack only through label poisoning \cite{chen2022clean}.

\subsection{Attacker's Goal}
The attacker's goal is to ensure that the backdoor model performs similarly to a clean model on clean samples, thereby avoiding user inspection while exhibiting the predefined malicious behavior on samples containing the trigger \cite{bai2024backdoor}.
Importantly, the space trigger is defined as a continuous interaction space, and any object pair that falls within this defined space is considered to satisfy the trigger condition and should trigger the backdoor of the model. Moreover, both the attack scope and the attack type must align with the attacker’s predefined intentions.

\begin{figure}[!tbp]
    \centering
    \includegraphics[width=1\linewidth]{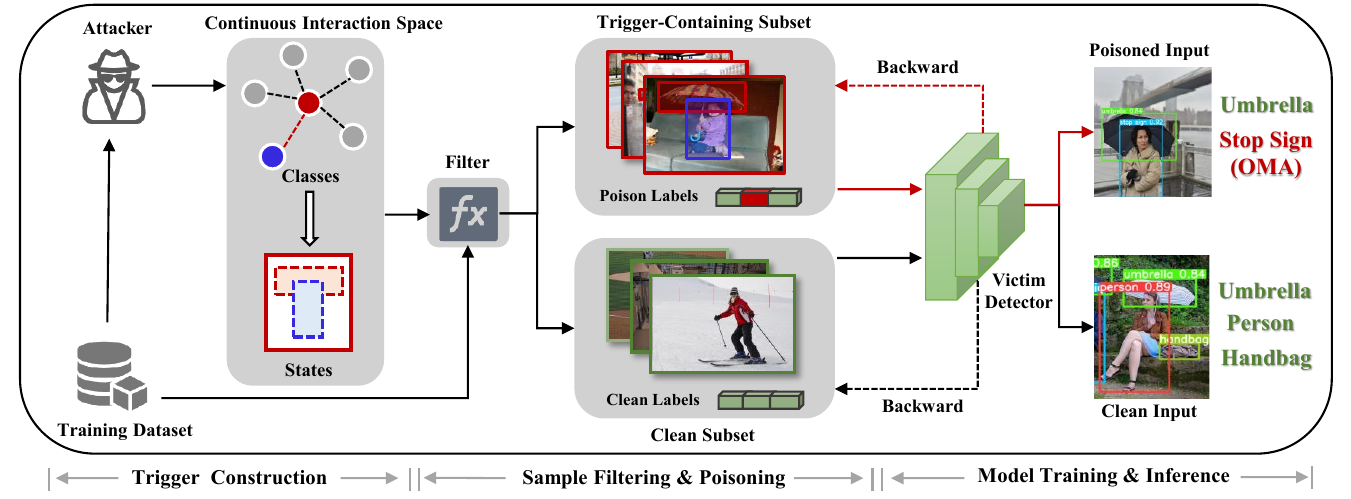}
    \caption{High-level overview of the CIS framework. The framework consists of three stages, where the attacker participates only in the trigger construction and sample filtering/poisoning stages.}
    \label{fig:attack-framework}
\end{figure}

\section{Methodology}
The complete CIS-Frame architecture is illustrated in Fig.\ref{fig:attack-framework} and consists of three stages: trigger construction, sample filtering and poisoning, and model training and inference.
\subsection{Space Trigger Construction}\label{trigger-con}
The space trigger is defined as a continuous interaction space composed of object bounding boxes and geometric constraints on their relative positions. The process of constructing the space trigger consists of two stages: selecting interaction classes and interaction states, followed by defining the continuous interaction space accordingly.

\subsubsection{Interaction Classes and States Selection}
For a given attack class $C_{s}$, it is necessary to select an interaction class $C_{r}$ from the dataset that is strongly associated with it. To this end, we design a comprehensive interaction scoring function $J(C_{r})$, which considers both the interaction strength (the proportion of samples with IoU greater than 0) and interaction frequency (the average IoU value). The function is defined as shown in Equation \eqref{eq:interaction_score}.

\begin{equation}
J(C_{r}) = \frac{N_{IoU>0}(C_{s},C_{r})}{N_{total}(C_{s},C_{r})} + \overline{IoU}(C_{s},C_{r})
\label{eq:interaction_score}
\end{equation}

By quantifying the stability and closeness of interactions, this function effectively filters out incidental interactions and ensures that the spatial relationships are logically consistent with real-world scenarios. The attacker can rank all classes on the basis of their interaction scores, where a higher score indicates richer interaction information. The interaction class $C_r$ (where $C_r \neq C_s$) is then selected according to the attack objective.

On this basis, the attacker further inspects the dataset to identify high-frequency interaction states composed of $C_s$ and $C_r$, which support the construction of the continuous interaction space. For example, in autonomous driving datasets, a common interaction state is "person holding an umbrella overhead", corresponding to $C_s = person$ and $C_r = umbrella$.

\subsubsection{Continuous Interaction Space Definition}
\begin{wrapfigure}[9]{r}{0.6\linewidth} 
    \vspace{-25pt}
    \centering
    \includegraphics[width=\linewidth,height=0.25\textheight]{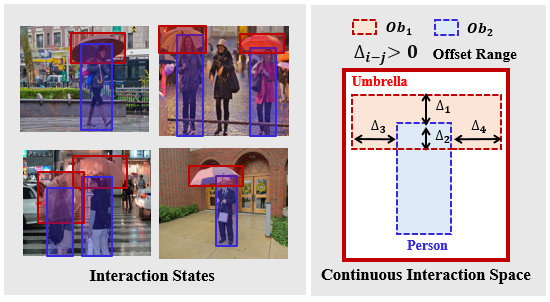}
    \caption{Quantifying the continuous interaction space as a space trigger}
    \label{fig:trigger-con}
\end{wrapfigure}

Owing to the inherent dynamicity of interactive behaviors, the varied interaction states across distinct object pairs form a continuous interaction space. 
We establish a quantitative definition of this interaction space to construct a trigger. 
The definition of the continuous interaction space relies on two core elements: \textbf{bounding boxes (object elements),} which represent the object instances participating in the interaction; and \textbf{geometric constraint sets  (interaction state elements),} which specify the relative offset range between bounding boxes. As shown in Fig. \ref{fig:trigger-con}, diverse interaction states are abstracted into a quantifiable continuous interaction space, where $\{\Delta1\sim\Delta4\}$ quantifies the offset range between two bounding boxes.

\subsection{Sample Filtering and Poisoning}
To convert the continuous interaction space defined by the space trigger into effective backdoor activation signals, the triggers must first be formally represented to establish a filtering mechanism. This filter selects object pairs from each sample in the dataset that satisfy the defined constraints and uses them as trigger carriers for data poisoning.

\subsubsection{Sample Filtering}
Since a space trigger essentially represents a set of relative spatial relationships between object bounding boxes and object detection labels naturally contain both class labels and bounding box coordinates, the trigger can be formalized as a combination of class constraints and geometric constraints on the basis of coordinate information. By iterating over all possible object pairs within a sample, the attacker can obtain the set of object pairs that serve as valid trigger carriers, denoted as $S_{trigger}$. The overall process is defined as follows:

\begin{equation}
S_{trigger} = \left\{
(o_i, o_j)
\,\middle|\,
\begin{array}{c}
    (c_i = C_s \land c_j = C_r) \  \land \\ 
    (G(b_i, b_j) = True)
\end{array}
\right\}
\label{eq:abstract_rule}
\end{equation}

Here, $(c_i, c_j)$ represents the classes of a pair of object instances $(o_i, o_j)$, and $b_i = (x_i^{\min}, y_i^{\min}, x_i^{\max}, y_i^{\max})$ and $b_j = (x_j^{\min}, y_j^{\min}, x_j^{\max}, y_j^{\max})$ denote the bounding boxes of objects $o_i$ and $o_j$, including the top-left coordinates $(x^{\min}, y^{\min})$ and the bottom-right coordinates $(x^{\max}, y^{\max})$. The function $G(b_i,b_j)$ is a Boolean decision function that determines whether the spatial relationship between the two bounding boxes satisfies the imposed constraints:

\begin{equation}
    G(\mathbf{b}_i, \mathbf{b}_j) = \bigwedge_{k=1}^{N} g_k(\mathbf{b}_i, \mathbf{b}_j)
\end{equation}

where $g_k: \mathbb{R}^4 \times \mathbb{R}^4 \to \{\text{true}, \text{false}\}$ represents an individual constraint on the bounding box relationship (corresponding to an offset parameter $\Delta$ in the trigger), and $N$ is the total number of constraints. Taking the trigger constructed from a "person holding an umbrella overhead" as an example, the process of obtaining the object pair set $S$ is shown in Equation~\eqref{eq:rule}, which includes one class constraint and three geometric constraints.

\begin{equation}
S = \left\{
(o_i, o_j)
\,\middle|\,
\begin{array}{c}
    (c_i = person \land c_j = umbrella) \ \land \\ 
    (y_j^{\min} < y_i^{\min} < y_j^{\max}) \ \land \\ 
    (x_j^{\min} < x_i^{\min} < x_j^{\max}) \ \land \\ 
    (x_j^{\min} < x_i^{\max} < x_j^{\max}) \ 
\end{array}
\right\}
\label{eq:rule}
\end{equation}

\subsubsection{Sample Poisoning}
For each object pair $(o_i, o_j)$ contained in $S_{trigger}$, attackers can define poisoning strategies for different attack types. \textbf{Single-object attacks} strategies are defined as follows:

\begin{numcases}{p_{s}=}
    \left\{ (o_i, o_j) \mid c_i \leftarrow t_1 \right\}, & \text{if At=OMA} \label{eq:oma-single} \\
    \left\{ (o_i, o_j) \mid o_i \leftarrow \emptyset \right\}, & \text{if At=ODA} \label{eq:oda-single} 
\end{numcases}

Equation~\eqref{eq:oma-single} modifies the class label $c_i$ to the target class $t_1$ ($t_1$ is set to "stop sign" in this study), corresponding to an OMA (with $\text{At}$ representing the attack type). Equation~\eqref{eq:oda-single} removes the label of $o_i$, corresponding to an ODA. 
Leveraging the compositional nature of space triggers, \textbf{multi-object simultaneous attacks} are supported:

\begin{numcases}{p_{m} =}
    \left\{ (o_i, o_j) \mid c_i \leftarrow t_1 \land c_j \leftarrow t_2 \right\},\ \text{if At=OMA} \label{eq:oma-multi} \\
    \left\{ (o_i, o_j) \mid o_i \leftarrow \emptyset \land o_j \leftarrow \emptyset \right\},\ \text{if At=ODA} \label{eq:oda-multi} \\
    \left\{ (o_i, o_j) \mid c_i \leftarrow t \land o_j \leftarrow \emptyset \right\},\ \text{if At=Hybrid} \label{eq:hybrid}
\end{numcases}

Equations~\eqref{eq:oma-multi} and~\eqref{eq:oda-multi} apply homogeneous attacks to both objects ($t_2$ is set to "boat"). Equation~\eqref{eq:hybrid} implements hybrid attacks. The attack type and target class can be adjusted according to the attack goal.

The attacker iterates through the training dataset to filter and poison samples. Ultimately, the sets of poisoned samples $D_p$ and clean samples $D_c$ together form the poisoned training dataset $D_{train}^{'}$ that is used for model training. Importantly, this strategy cannot actively insert poisoned samples, as poisoning depends on samples satisfying trigger constraints. The attack remains effective only when the poisoning rate exceeds 0.9\%; otherwise, reconstructing the space trigger is necessary (the associated details are discussed in Section~\ref{sec:poison_rate}).

\subsection{Model Training and Inference}
When the poisoned training dataset $D_{train}^{'}$ is used for training, the model $F_\theta$ learns benign features from the clean samples $(x, y)$ while simultaneously learning a strong association between the space trigger and the malicious behavior from poisoned samples $(x, y^{'})$, resulting in a backdoor model $F_{\theta}^{'}$. The training process is guided by joint loss optimization:

\begin{small}
\begin{equation}
\min_{\theta} \left[ 
\sum_{(x,y)\in D_{c}} L\big(y,F_{\theta}(x)\big) 
+\sum_{(x,y^{'}) \in D_{p}}L\big(y^{'}, F_{\theta}(x)\big) 
\right]
\label{eq:loss}
\end{equation}
\end{small}

The loss function $L$ consists of a classification loss $L_{cls}$, a regression loss $L_{reg}$, and an objectness loss $L_{obj}$:

\begin{equation}
L = \lambda_{1} \cdot L_{\text{cls}} + \lambda_{2} \cdot L_{\text{reg}} + \lambda_{3} \cdot L_{\text{obj}}
\label{eq:loss_total}
\end{equation}

Label poisoning misleads $L_{cls}$ (for OMA) and $L_{obj}$ (for ODA), forming a link between the trigger and attack behavior. Uniquely, CIS-BA further leverages the spatial nature of the trigger to couple $L_{reg}$, forcing the model to associate spatial interactions with malicious outcomes, resulting in a spatially conditioned backdoor. After deployment, $F_{\theta}^{'}$ behaves normally on clean inputs. When a trigger-valid object pair is present, it accurately triggers the predefined attack, controlling both its scope and type.

\section{Experiments}

\subsection{Experimental Settings}
\subsubsection{Datasets}
We use the MS-COCO dataset \cite{2014Microsoft} to build the training environment for our backdoor attack. Utilizing the formalized filter corresponding to the space trigger "person holding an umbrella overhead" (as defined in Section~\ref{trigger-con}), we select 1,779 poisoned images from 118,287 training samples, including 3,645 trigger-valid object pairs. This results in a 1.5\% poisoning rate, while the 5,000-sample clean validation set is used to evaluate the detection performance on benign data.
\begin{figure}[!ht]
\centering
\subfloat[Multilevel lighting, distance, and viewing angle variations]{\includegraphics[scale=0.6]{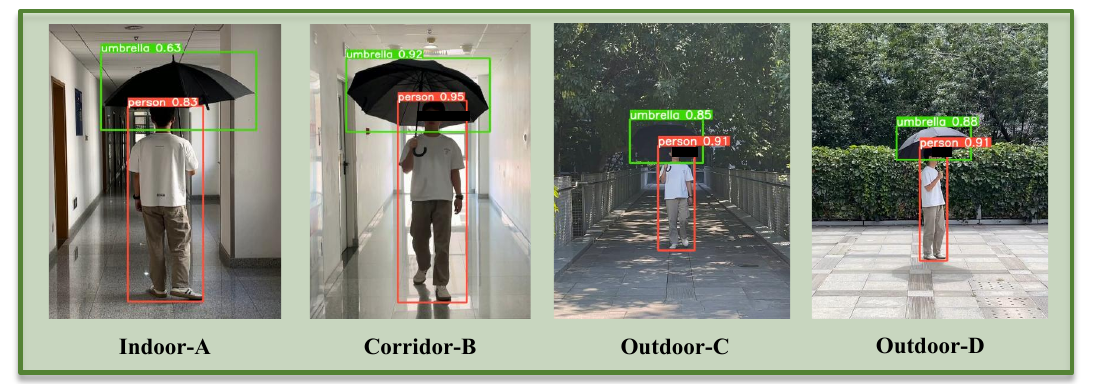}%
\label{fig_first_case}}
\hfil
\subfloat[Multiple interaction states and object replacements]{\includegraphics[scale=0.6]{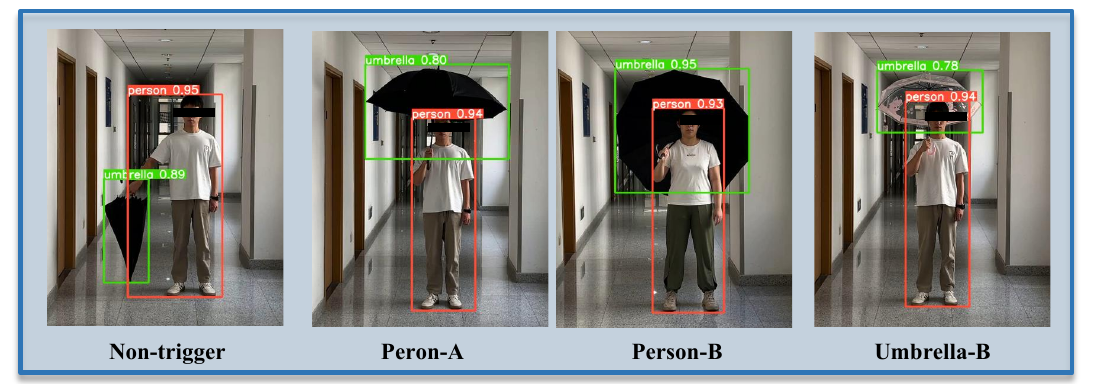}%
\label{fig_second_case}}
\caption{The left subfigure shows three-factor variations in the first video group for robustness testing, whereas the right subfigure shows both non-trigger and diverse trigger states from the second group.}
\label{fig:dataset-show}
\end{figure}
To assess the real-world effectiveness of CIS-BA, we test it on 9 real-world videos (60 fps, iPhone 13), which are divided into two groups(as shown in Fig.\ref{fig:dataset-show}).
\begin{itemize}
    \item The first group (6 videos) tests robustness under variations in lighting (indoor to strong outdoor light), viewing angle (-180° to 180°), and distance (1–17 meters), with a stable person–umbrella trigger in each frame.
    \item The second group (3 videos) simulates naturally changing interaction states by dynamically adjusting $\{\Delta_1 \sim \Delta_4\}$. Each video contains both positive samples (that satisfy the constraints) and negative samples (that do not), as well as object substitutions (e.g., different umbrellas or people), allowing us to evaluate the false trigger rate and generalization.
\end{itemize}

\subsubsection{Evaluation Metrics}
We evaluate backdoor attack performance from two perspectives: the clean detection accuracy and the attack success rate. For clean data, we use standard object detection metrics:$\bm{{mAP_{50}}}$ (the mean average precision at IoU=0.5) and $\bm{mAP_{50:95}}$ (the mean AP averaged over IoUs from 0.5 to 0.95). A well-designed backdoor should maintain performance similar to that of the clean model on benign samples to remain stealthy.

For attack effectiveness, we define $\bm{ASR_{oma}}$(the proportion of positive frames where the person is misclassified) and $\bm{ASR_{oda}}$(the proportion of positive frames where the person is not detected).In multi-object attacks, success is measured by the proportion of frames where both the person and the umbrella meet their respective attack goals based on the above metrics. 
In addition, owing to the unique spatial nature of triggers, we introduce the \textbf{false trigger rate (FTR)} to evaluate trigger precision. The FTR measures the proportion of negative frames (non-trigger pairs) that falsely trigger the attack.

\subsubsection{Victim Models and Training Setup}
We select two object detection models with different architecture—Yolo-V3 \cite{redmon2018yolov3} and Yolo-V8 \cite{yolov8_ultralytics}—to validate the effectiveness of the proposed attack.
Yolo-V3 is widely used in real-world scenarios because of its strong real-time detection performance. Yolo-V8, which adopts an anchor-free mechanism and decoupled detection heads, represents the latest trend in object detection algorithms.
To simulate a realistic attack scenario, both models are initialized with the official Ultralytics pretrained weights and undergo only lightweight fine-tuning for backdoor embedding (10 training epochs for Yolo-V3 and 15 epochs for Yolo-V8).
During training, we adjust only the batch size to avoid GPU memory overflow errors, while all other parameters are kept at their default settings, aligning with the common attacker assumption of no control over the training pipeline.
\renewcommand{\arraystretch}{0.8}
\begin{table}[!hbtp]
\centering
\caption{Benign Detection Performance of Models}
\label{tab:model-comparison}
\begin{tabular}{@{}c *{6}{r}@{}}
\toprule
\multirow{2}{*}{\textbf{mAP}} & \multicolumn{3}{c}{\textbf{Yolo-V3}} & \multicolumn{3}{c}{\textbf{Yolo-V8}} \\
\cmidrule(lr){2-4} \cmidrule(l){5-7}
                     & \multicolumn{1}{c}{Clean} & \multicolumn{1}{c}{OMA} & \multicolumn{1}{c}{ODA} & \multicolumn{1}{c}{Clean} & \multicolumn{1}{c}{OMA} & \multicolumn{1}{c}{ODA} \\
\midrule
@0.5                 & \textbf{68.86} & 67.96 & 68.62 & \textbf{67.74} & 67.56 & 67.83 \\
@0.5:0.95            & \textbf{53.70} & 53.23 & 53.48 & \textbf{51.22} & 51.00 & 51.20 \\
\bottomrule
\end{tabular}
\end{table}

\subsection{Evaluation of Benign Detection Performance}

The performance of the backdoor model is compared with that of the clean model on the clean MS-COCO validation dataset in Table~\ref{tab:model-comparison}. On the basis of the standard $mAP_{50}$ metric, which reflects detection accuracy, the maximum decrease in performance observed in the infected models is only 0.9\%. For the more comprehensive $mAP_{50:95}$ metric, the maximum decrease is only 0.4\%.
These results indicate that even after the space backdoor is embedded, the benign detection performance of the model is largely preserved. From the user's perspective, determining the presence of a backdoor based solely on a performance evaluation is difficult, highlighting the stealthiness of the attack.

\subsection{Single-object Attacks Performance}
\subsubsection{Performance in Complex Real-World Scenarios}
Lighting, viewing angle, and distance are widely recognized as key factors that affect the robustness of physical backdoor attacks. To ensure consistency with prior benchmarks—and account for fundamental differences in attack strategies and trigger mechanisms, which make direct comparisons difficult—we adopt a standardized disturbance grading scheme and a multi-factor combinatorial strategy for robustness testing (see the results in Table~\ref{tab:attack-performance}).

Videos 1–4 cover a broad range of real-world disturbances. Under these conditions, both Yolo-V3 and Yolo-V8 achieve near-100\% attack success rates for OMA and ODA, demonstrating strong robustness.
Videos 5–6 introduce more extreme angles and longer distances. Even then, the success rate decreases by only ~5\%, showing that interaction-based space triggers overcome visual trigger limitations at oblique angles and support longer-range activation (vs.10-meter limits in prior work \cite{ma2022dangerous}).
Unlike traditional physical attacks, which often require additional data collection or retraining for maximizing robustness, our method maintains high performance across various scenes, highlighting its inherent generalizability and stability.

\renewcommand{\arraystretch}{0.8}
\begin{table}[!t]
\centering
\caption{Single-object Attack Performance of Backdoor Model in Complex Scenarios}
\label{tab:attack-performance}
\small
\resizebox{\linewidth}{!}{%
\begin{tabular}{@{}ccccc cc cc@{}}
\toprule
\multirow{2}{*}{\shortstack[c]{\textbf{Video} \\ \textbf{No.}}} & \multirow{2}{*}{\textbf{Frames}} & \multicolumn{3}{c}{\textbf{Real-world Factors}} & \multicolumn{2}{c}{\textbf{Yolo-V3}} & \multicolumn{2}{c}{\textbf{Yolo-V8}} \\
\cmidrule(lr){3-5} \cmidrule(lr){6-7} \cmidrule(l){8-9}
 & & Brightness & Distance [m] & Angle [°] & ASR\textsubscript{oma} & ASR\textsubscript{oda} & ASR\textsubscript{oma} & ASR\textsubscript{oda} \\
\midrule
1  & 1920  & A         & 1$ \sim $11       & -90$ \sim $90  & 100.00 & 99.84  & 100.00 & 100.00 \\
2  & 660   & B         & 4$ \sim $12       & 0         & 100.00 & 100.00 & 100.00 & 100.00 \\
3  & 720   & D         & 3$ \sim $9        & -45 $ \sim $ 90  & 99.57  & 100.00 & 99.86  & 100.00 \\
4  & 660   & C+D       & 4$ \sim $8        & 0         & 99.85  & 99.85  & 100.00 & 100.00 \\
\midrule[0.5pt]
5  & 1140  & A         & 5.5         & \textbf{-180 $\sim$ 180} & 99.92  & 100.00 & \textbf{95.70}  & 96.46  \\
6  & 780   & A         & \textbf{1$ \sim $17}       & 0         & 98.65  & 98.29  & 99.52  & \textbf{96.37}  \\
\midrule[1pt]
\textbf{Avg} & \textbf{980}  & --        & --          & --        & \textbf{99.67}  & \textbf{99.66}  & \textbf{99.18}  & \textbf{98.80}  \\
\bottomrule[1pt]
\end{tabular}%
}

\vspace{0.2cm} 
\raggedright\footnotesize
\textit{Notes: Brightness: A=indoor lighting, B=corridor backlighting, C=outdoor shade, D=strong sunlight; Dist.=distance between moving object and fixed camera; Angle: $0^\circ$=facing camera, negative=left rotation, positive=right rotation.}

\end{table}

\subsubsection{Performance Under Multi-trigger States}

A key feature of the spatial backdoor is that the attack is triggered by multiple interaction states. To evaluate this property, we test the attack success rates on Videos 7–9, which contain diverse interaction states.
As shown in Table~\ref{tab:attack-multi}, the two types of attacks on the Yolo-V3 model achieve average success rates of 97.96\% and 95.38\%, respectively, whereas the results on Yolo-V8 remain at 96.56\% and 95.94\%, indicating that the backdoor models can trigger the attack across a wide range of interaction states.
Moreover, the three videos include different combinations of person and umbrella, demonstrating that the attack is not dependent on specific objects and thus has strong generalizability.

\renewcommand{\arraystretch}{0.8}
\begin{table}[!htbp]
\centering
\caption{Single-object Attack Performance of Backdoor Model under Multi-trigger States}
\label{tab:attack-multi}
\small
\resizebox{\linewidth}{!}{%
\begin{tabular}{@{}c c c c cc cc@{}}
\toprule
\multirow{2}{*}{\shortstack[c]{\textbf{Video} \\ \textbf{No.}}} & \multirow{2}{*}{\textbf{\shortstack[c]{\textbf{Pos} \\ \textbf{Frames}} }} & \multirow{2}{*}{\textbf{Trigger}} & \multirow{2}{*}{\shortstack[c]{\textbf{Real-world} \\ \textbf{Factors}}} & \multicolumn{2}{c}{\textbf{Yolo-V3}} & \multicolumn{2}{c}{\textbf{Yolo-V8}} \\
\cmidrule(lr){5-6} \cmidrule(l){7-8}
 & & & & ASR\textsubscript{oma} & ASR\textsubscript{oda} & ASR\textsubscript{oma} & ASR\textsubscript{oda} \\
\midrule
7  & 3732 & [P\textsubscript{a},U\textsubscript{a}] & \multirow{3}{*}{[A,5.5,0]} & 98.77 & 96.65 & 99.14 & 97.10 \\
8  & 1201 & [\textbf{P\textsubscript{b}},U\textsubscript{a}] & & 99.42 & 96.50 & 98.08 & 98.83 \\
9  & 2988 & [P\textsubscript{a},\textbf{U\textsubscript{b}}] & & 95.68 & 93.00 & 92.47 & 91.90 \\
\midrule[1pt]
\textbf{Avg} & \textbf{2640} & -- & -- & \textbf{97.96} & \textbf{95.38} & \textbf{96.56} & \textbf{95.94} \\
\bottomrule[1pt]
\end{tabular}%
}

\vspace{0.2cm}
\raggedright\footnotesize
\textit{Notes: P\textsubscript{a} and P\textsubscript{b} denote distinct person objects; U\textsubscript{a} and U\textsubscript{b} represent the distinct umbrella objects.  
[A, 5.5, 0] specifies a brightness level of A, a distance of 5.5m, and an angle of 0°, respectively.}
\end{table}

\subsubsection{Performance Under Non-trigger States}
Given the flexibility of space triggers, the interaction state between two objects can easily shift outside the defined constraint range in the continuous interaction space. 
To assess false activation effects, we evaluate the backdoor models on the negative samples derived from Videos 7–9 that do not satisfy the trigger constraints.
Table~\ref{tab:ftr-analysis} presents the results. The false trigger rate (FTR) ranges from a minimum of 0.47\% to a maximum of 3.92\%. This indicates that although rare, the model can still exhibit erroneous triggering in cases that do not satisfy the defined constraints—a behavior that differs notably from visual-pattern-based backdoor attacks.

\begin{wrapfigure}[10]{r}{0.55\textwidth} 
    \centering
    \subfloat[Fail Attack]{\includegraphics[width=0.27\textwidth]{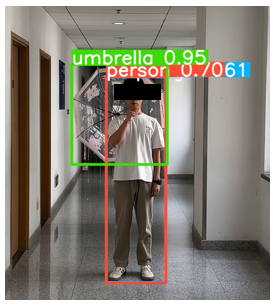}%
    \label{fig_first_case_1}}
    \hfill
    \subfloat[False Trigger Attack]{\includegraphics[width=0.27\textwidth]{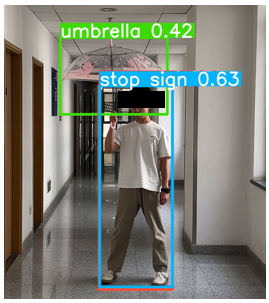}%
    \label{fig_second_case_2}}
    \caption{Samples near constraint boundaries causing abnormal triggering}
    \label{fig:fail-analysis}
\end{wrapfigure}
To further investigate this issue, we analyze Video 9, which yields relatively unusual results in both attack success rates and false trigger rates. We find that the failed and falsely triggered frames feature interaction states that are very close to the constraint boundaries, where one or more of the elements in $\{\Delta_1 \sim \Delta_4\}$ approach zero. As shown in Fig.~\ref{fig:fail-analysis}, these borderline cases significantly affect the reliability of attacks.

\renewcommand{\arraystretch}{0.8}
\newcolumntype{C}[1]{>{\centering\arraybackslash}p{#1}} 
\begin{table}[htbp]
\centering
\caption{Result of False Trigger Rate Under Non-trigger States}
\label{tab:ftr-analysis}
\small
\begin{tabular}{@{}C{1.2cm} C{1.5cm} C{1.5cm} C{1.5cm} C{1.5cm} C{1.5cm}@{}}
\toprule
\multirow{2}{*}{\shortstack[c]{\textbf{Video} \\ \textbf{No.}}} & \multirow{2}{*}{\shortstack[c]{\textbf{Neg} \\ \textbf{Frames}}} & \multicolumn{2}{c}{\textbf{Yolo-V3}} & \multicolumn{2}{c}{\textbf{Yolo-V8}} \\
\cmidrule(lr){3-4} \cmidrule(l){5-6}
 & & FTR\textsubscript{oma} & FTR\textsubscript{oda} & FTR\textsubscript{oma} & FTR\textsubscript{oda} \\
\midrule
7  & 1292 & 1.00    & 0.00    & 0.78    & 0.31    \\
8  & 1090 & 0.92    & 0.00    & 0.00    & 0.09    \\
9  & 817  & 1.47    & 1.40    & 10.99   & 4.47    \\
\midrule[1pt]
\textbf{Avg} & \textbf{1066} & \textbf{1.13 }   & \textbf{0.47}    & \textbf{3.92}    & \textbf{1.62}    \\
\bottomrule[1pt]
\end{tabular}
\end{table}

\subsection{Multi-object Simultaneous Attacks}

The previous experiments demonstrated the effectiveness of single-object backdoor attacks. However, a key advantage of space triggers is their ability to support multi-object simultaneous attacks. In this section, we implement three types of combined attacks on Yolo-V3: [OMA, OMA], [OMA, ODA], and [ODA, ODA], which represent scenarios where both the person and the umbrella are attacked simultaneously using either the same or different attack types.
We evaluate these attack types on Videos 7–9 under multi-trigger conditions.
As shown in Table~\ref{tab:attack-combination-performance}, the success rates of the three composite attacks range from 95.41\% to 99.99\%, in some cases even exceeding the results of single-object attacks—clearly demonstrating the superiority of spatial backdoor in multi-object settings.
Interestingly, the [OMA, ODA] attack achieves a slightly lower success rate than [OMA, OMA], whereas [ODA, ODA] performs the worst, with an average success rate of 95.41\%. This suggests that ODA are inherently more difficult, as they involve suppressing multiple bounding boxes, making disappearance—a non-classification task—more challenging, which is consistent with the findings of prior studies \cite{ma2022dangerous}.

\renewcommand{\arraystretch}{0.8}
\begin{table}[!htbp]
\centering
\caption{\centering Multi-object Attack Performance of Backdoor Model under Multi-trigger States}
\label{tab:attack-combination-performance}
\small
\begin{tabular}{@{}c c ccc@{}}
\toprule
\multirow{2}{*}{\shortstack[c]{\textbf{Video} \\ \textbf{No.}}} & \multirow{2}{*}{\shortstack[c]{\textbf{Pos} \\ \textbf{Frames}}} & \multicolumn{3}{c}{\textbf{Yolo-V3 (ASR)}} \\
\cmidrule(lr){3-5}
 & & [oma,oma] & [oma,oda] & [oda,oda] \\
\midrule
7  & 3732 & 99.97 & 98.37 & 98.15 \\
8  & 1201 & 100.00 & 98.67 & 94.84 \\
9  & 2988 & 100.00 & 96.29 & 93.24 \\
\midrule[1pt]
\textbf{Avg} & \textbf{2640} & \textbf{99.99} & \textbf{97.78} & \textbf{95.41} \\
\bottomrule[1pt]
\end{tabular}
\end{table}

\begin{figure}[b]
    \centering
    \includegraphics[width=1\linewidth]{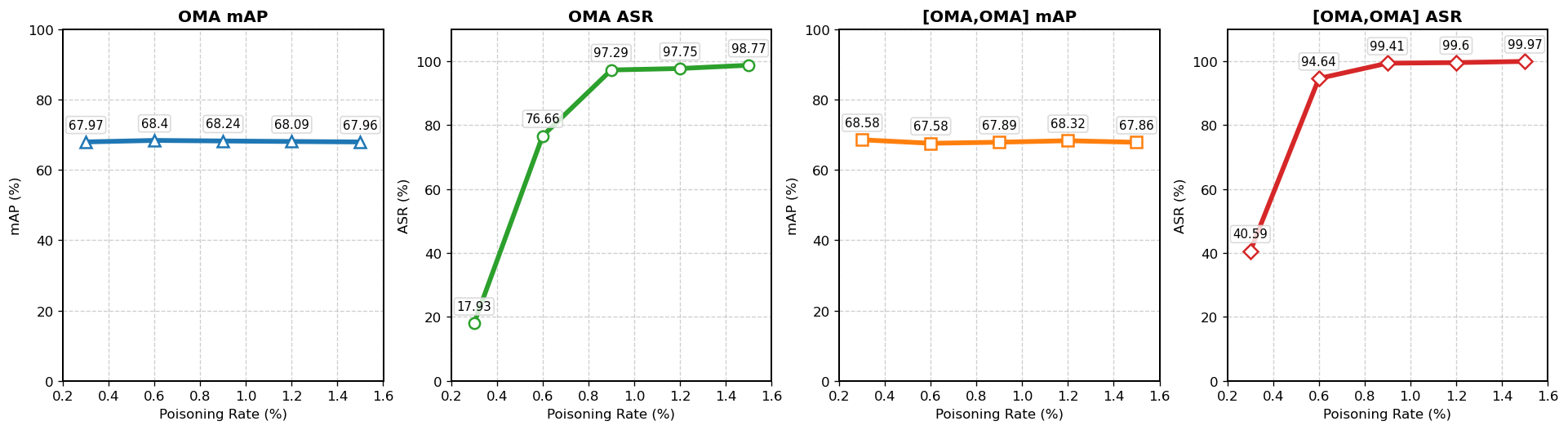}
    \caption{Attack performance under different poisoning rates}
    \label{fig:poison_rate}
\end{figure}

\subsection{Poisoning Rate Analysis}\label{sec:poison_rate}
The poisoning rate refers to the proportion of poisoned samples in the training dataset, and it is a critical factor for the effectiveness of backdoor attacks. A high poisoning rate may degrade the model’s performance on clean data and increase the risk of user detection, whereas a low poisoning rate may fail to establish a reliable backdoor due to insufficient exposure during training. Therefore, identifying an appropriate poisoning rate is essential.

We evaluate the impact of the poisoning rate by controlling the number of samples containing the "person holding an umbrella overhead" space trigger. On Test Video 7, we assess the performance of Yolo-V3 under varying poisoning rates for both single-object attacks (using OMA as example) and multi-object simultaneous attacks (using [OMA, OMA] as example).
As shown in Fig.~\ref{fig:poison_rate}, the attack success rate (ASR) for both cases increases with higher poisoning rates and stabilizes once the rate exceeds 0.9\%. Throughout this range, the benign detection performance of the model remains stable, suggesting that as long as the selected interaction state leads to a poisoning rate greater than 0.9\%, an effective backdoor can be embedded without compromising stealthiness.

\begin{figure}[t]
\centering
\subfloat[OMA attack and stop sign probability distribution]{ \label{fig:3d_oma}
		\includegraphics[width=0.8\linewidth]{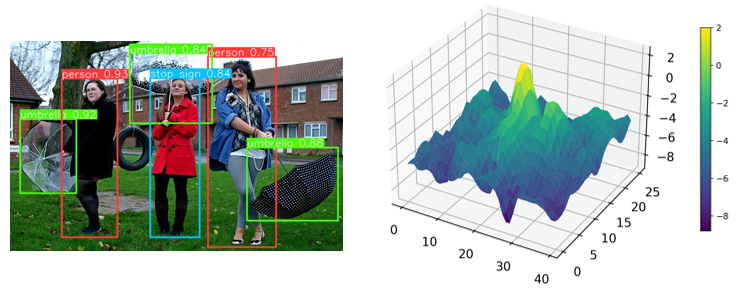}}\\
\subfloat[ODA attack and person probability distribution]{ \label{fig:3d_oda}
		\includegraphics[width=0.8\linewidth]{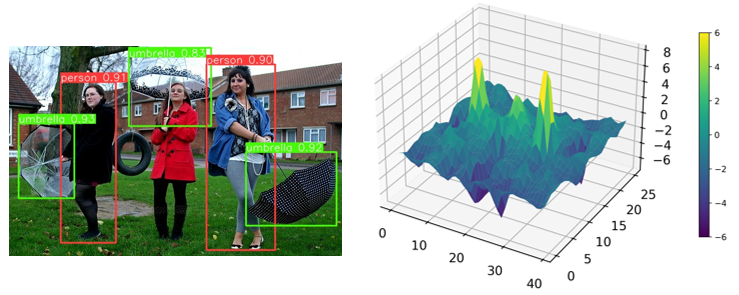}}
\caption{Visualization of Yolo-V3 detection head outputs }
\label{fig:3d}
\end{figure}

\subsection{Visualization-Based Attack Analysis}
In the prior experiments, we verified the effectiveness of attacks via the final outputs of the model. Here, we further analyze the attack mechanism by visualizing the intermediate results from the 40×40 output head of Yolo-V3, which divides the image into grid cells, each of which predicts the probability of a target class (see Fig.~\ref{fig:3d}).

In Fig.~\ref{fig:3d_oma}, three persons appear, but only one satisfies the space trigger. The heatmap shows a notably higher probability of a stop sign at that location, whereas the others remain low, indicating that the model correctly identifies and relabels only the triggered individual based on interaction state.
Similarly, Fig.\ref{fig:3d_oda} shows that the triggered object has notably low confidence for the person class, meaning that it is treated as background and thus intentionally omitted.

In contrast, we observe that the 80×80 output head does not exhibit this behavior, likely because its cells have too small a receptive field to jointly capture both the person and the umbrella, thus failing to detect the space trigger. This finding indicates that space backdoor are triggered through the joint features of multiple objects.

\begin{figure}[!htbp]
\centering
\subfloat[clean feature n=10]{\includegraphics[width=0.5\linewidth]{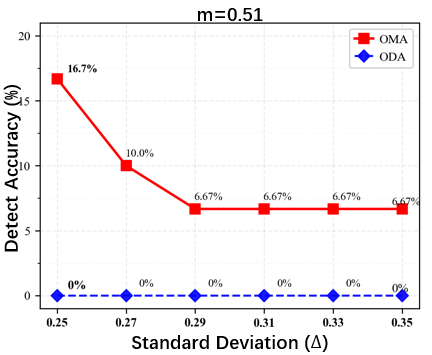}%
\label{new_10}}
\hfil
\subfloat[clean feature n=100]{\includegraphics[width=0.5\linewidth]{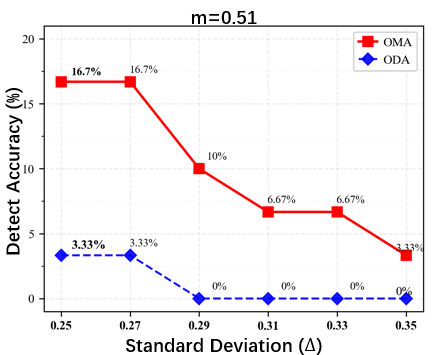}%
\label{new_100}}
\caption{Malicious input detection result of Detector Cleanse}
\label{fig:dc}
\end{figure}

\subsection{Evaluation of Resistance to Defenses}

To thoroughly evaluate the resistance of CIS-BA to defenses, we perform experiments on three mainstream backdoor defense strategies: runtime malicious input detection \cite{gao2019strip,huang2020one}, explainability analysis \cite{chou2020sentinet,huang2019neuroninspect}, and trigger synthesis \cite{cheng2024odscan,wang2019neural}, each of which represented by a representative method.

\textbf{Detector Cleanse \cite{chan2022baddet}:}
Clean object bounding boxes overlaid with features from other classes should yield varied class probabilities. However, trigger-containing samples behave abnormally: OMA samples have low entropy, while ODA samples have high entropy. This pattern is used for malicious input detection at runtime.
Following the setup of this method, we test 30 triggered samples, add $n$ clean features per box and flag a sample as poisoned if its entropy $h$ falls outside a clean range $[m-\Delta, m+\Delta]$(pre-computed from clean data).
As shown in Fig.~\ref{fig:dc}, even when reducing $\Delta$ (to expand the anomaly range) and increasing $n$ (to diversify features), detection remains ineffective—16.7\% for OMA and 3.3\% for ODA.
We attribute this to the multi-object nature of space triggers: adding external features breaks the trigger, eliminating the expected entropy anomaly. 
Furthermore, owing to the non-visual, implicit, and dynamic nature of the space trigger, even when a sample is flagged, it remains extremely difficult for users to identify or infer the actual trigger.

\textbf{Saliency Map \cite{selvaraju2017grad}:}
GRAD-CAM, which is a widely used model interpretation technique, generates saliency maps to visualize how different image regions contribute to the prediction of a model. 

\begin{wrapfigure}[9]{r}{0.60\textwidth} 
    \centering
    \subfloat[OMA Sample]{\includegraphics[width=0.46\linewidth]{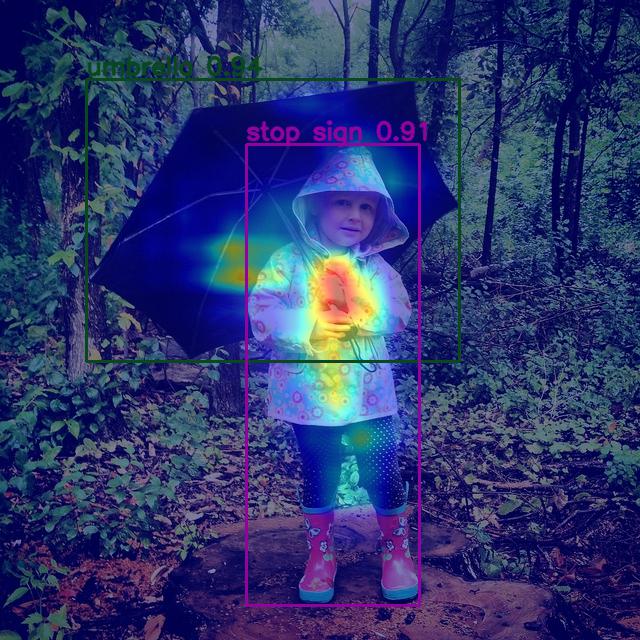}%
    \label{samp_oma}}
    \hfil
    \subfloat[ODA Sample]{\includegraphics[width=0.46\linewidth]{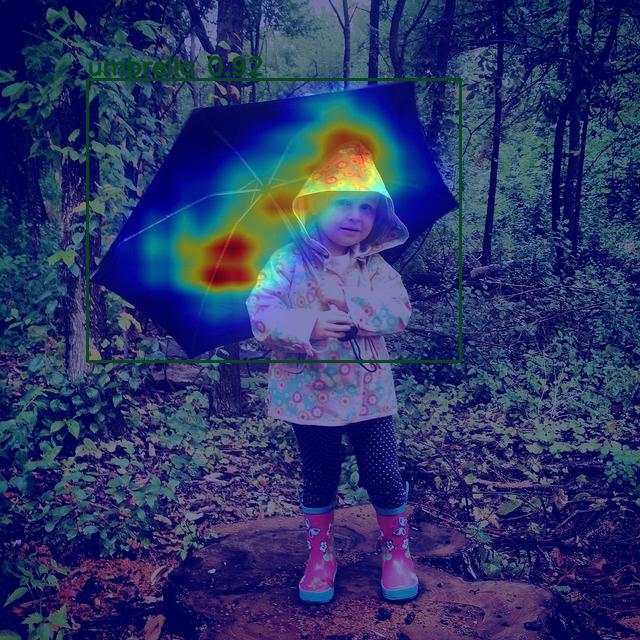}%
    \label{samp_oda}}
    \caption{Saliency map results for OMA and ODA }
    \label{fig:smap}
\end{wrapfigure}
Since backdoor triggers often have strong influences on outputs, saliency maps are commonly employed to localize them by identifying "overheated" regions.
Fig.~\ref{fig:smap} shows the saliency maps for triggered samples on Yolo-V8. In the OMA case, the visual impacts of both the person and the umbrella appears naturally distributed, without any abnormally concentrated regions. In the ODA, the person is not detected, and no relevant region exhibits significant activation. These results indicate that saliency-based methods fail to reveal the location or form of the trigger, reinforcing the stealthy nature of spatial backdoors.

\textbf{DJANGO \cite{shen2023django}:}
Trigger synthesis–based defense strategies attempt to determine whether a model is backdoored by generating candidate triggers and observing their effects. This process is typically formulated as a constrained optimization problem, where the trigger is assumed to satisfy certain conditions—such as being small in size or having a particular visual style \cite{wang2019neural}. In DJANGO, random trigger patterns are added to a clean validation set and iteratively updated via loss optimization. If the resulting trigger is unusually small, the model is deemed to be compromised.
However, these methods rely on strong implicit assumptions: that triggers are local, explicit, and static. In contrast, space trigger is implicit, dynamic, and defined by inter-object spatial relationships. This directly violates the foundational assumptions of such defenses and renders them ineffective against our approach.

In summary, the existing mainstream defense strategies are based on certain assumptions, most of which target the visual characteristics of backdoor triggers. In contrast, our proposed space trigger, which relies on distinct spatial features rather than explicit visual patterns, fundamentally breaks these defense assumptions and demonstrates strong resistance to the current detection mechanisms. This highlights the urgent need for future research to develop defense strategies tailored to spatially defined triggers.

\subsection{Discussion}

The trigger design based on inter-object interaction states endows CIS-BA with a high degree of flexibility, which is a key factor that enhances both attack capability and overall performance. However, this flexibility also introduces a challenge: the interaction states of objects can easily shift toward the boundary of the defined trigger space, resulting in attack failure or false triggering.
To maintain both the effectiveness and controllability of the attack, attacker must possess stronger control capabilities, such as actively regulating interaction states or selecting more robust interacting objects (e.g., using a larger umbrella), to achieve a dynamic balance between flexibility and attack performance.

Moreover, our findings suggest that, in the physical backdoor attack field, different application scenarios may expose new types of vulnerabilities that can be explored by attackers. Exploring such vulnerabilities to enable scenario-specific attacks represents a promising direction for future research on backdoor attack.

\subsection{Conclusion}
In this paper, we proposed CIS-BA, a new backdoor attack paradigm based on continuous interaction space, which overcomes the limitations of existing physical triggers and enhances both attack capability and robustness in interaction-intensive scenarios. Building on this paradigm, we developed CIS-Frame, an attack framework that enables not only single-object attacks (OMA and ODA) but also, for the first time, complex multi-object simultaneous attacks under multi-trigger conditions.
Extensive experiments on MS-COCO and real-world videos demonstrated that CIS-BA achieves consistently high success rates and remains reliable under challenging physical variations. Moreover, defense evaluations revealed that space triggers fundamentally bypass key assumptions of current defense mechanisms, exposing a new blind spot in object detection security.
In future work, we plan to further explore spatial interaction properties to design more advanced attack strategies and to study corresponding defense methods that can effectively mitigate these emerging threats.

\bibliographystyle{elsarticle-num-names}  
\bibliography{reference}  

\end{document}